%% file: main.tex
\documentclass[runningheads]{llncs}

\usepackage{eccv}

\input{preamble}

\usepackage{eccvabbrv}

\usepackage{graphicx}
\usepackage{booktabs}
\usepackage{marvosym}

\usepackage[accsupp]{axessibility}  

\usepackage{hyperref}

\begin{document}

\title{\sysname{}: Compact Gaussian Splatting via Human-Centric Prediction}


\makeatletter
\renewcommand{\@fnsymbol}[1]{\ifcase#1\or\Letter\else\@arabic{#1}\fi}
\makeatother

\author{Yujie Guo\inst{1} \and
Yudong Jin\inst{1} \and
Lingteng Qiu\inst{3} \and
Zehong Shen\inst{1} \and
Zhen Xu\inst{1} \and
Jing Zhang\inst{2} \and
Xianchao Shen\inst{2} \and
Hujun Bao\inst{1} \and
Sida Peng\inst{1} \and
Xiaowei Zhou\inst{1}\thanks{Corresponding author.}}

\authorrunning{Y.~Guo et al.}

\institute{State Key Lab of CAD\&CG, Zhejiang University, Hangzhou, China \and
ByteDance, Beijing, China \and
The Chinese University of Hong Kong, Shenzhen, China}

\maketitle

\input{sec/0_abstract}
\input{sec/1_intro}
\input{sec/2_related_work}
\input{sec/3_method}
\input{sec/4_experiments}
\input{sec/5_conclusion}

\input{sec/acknowledgements}

\bibliographystyle{splncs04}
\bibliography{main}

\startsupplementarymaterial
\begin{center}
    {\Large\bfseries Supplementary Material\par}
\end{center}
\input{sec/X_suppl}

\end{document}

%% file: preamble.tex
\newcommand{\PAR}[1]{\vskip4pt \noindent{\bf #1~}}

\newcommand{\sysname}[0]{PointSplat}
\newcommand{\startsupplementarymaterial}{%
    \clearpage
    \appendix
    \renewcommand{\theHsection}{appendix.\thesection}%
}

\usepackage{multirow}
\usepackage{float}
\usepackage{placeins}
\usepackage{algorithm}
\usepackage{algpseudocode}

%% file: sec/0_abstract.tex
\begin{abstract}

Producing 3D human representations from input views on the fly is essential for immersive live streaming systems, where representation compactness is as critical as high fidelity given limited computational power and transmission bandwidth. Although recent feed-forward reconstruction methods achieve impressive quality through the view-centric prediction of 3D representations, they repeatedly encode the same subject content across multiple views, leading to significant inter-view redundancy. Our key insight is to perform predictions directly in 3D space, enabling the network to learn and produce a highly compact representation. To this end, we propose PointSplat, a novel human-centric approach that directly infers Gaussian primitives from an input point set. The proposed method first estimates a coarse geometric proxy and performs ray casting to prune redundant points and establish explicit 2D--3D correspondences. Subsequently, it employs a Point-Image Transformer to fuse appearance and geometry features, predicting Gaussian attributes in a single forward pass. This design restricts predictions to foreground regions of interest, substantially reducing the total number of Gaussians while improving novel-view rendering quality. Extensive experiments demonstrate that PointSplat achieves higher efficiency and quality while exhibiting strong robustness to variations in view count and image resolution across multiple datasets. The project page is available at \url{https://zju3dv.github.io/pointsplat}.

\keywords{3D Representation \and Compactness \and Feed--Forward Model}


\end{abstract}

%% file: sec/1_intro.tex
\section{Introduction}
\label{sec:intro}

This paper addresses the problem of producing compact 3D human representations from input views, which is critical for immersive live streaming systems, holographic communication, and related applications.
These real-time scenarios demand high-fidelity and high-speed reconstruction, while also requiring low hardware cost and efficient transmission to deliver an immersive experience to users~\cite{Lee_Tabatabai_Tashiro_2015,Mildenhall2021NeRF,levoy1996light,gortler1996lumigraph,Muller2022InstantNGH}.
Traditional light-field-based approaches~\cite{levoy1996light,gortler1996lumigraph,ng2005light,mildenhall2019local},
which employ dense camera arrays for view interpolation, can achieve high-quality rendering with low latency.
However, their reliance on costly capture hardware severely limits scalability and deployment in practical settings.

\begin{figure*}[t]
    \centering
    \includegraphics[width=0.95\textwidth]{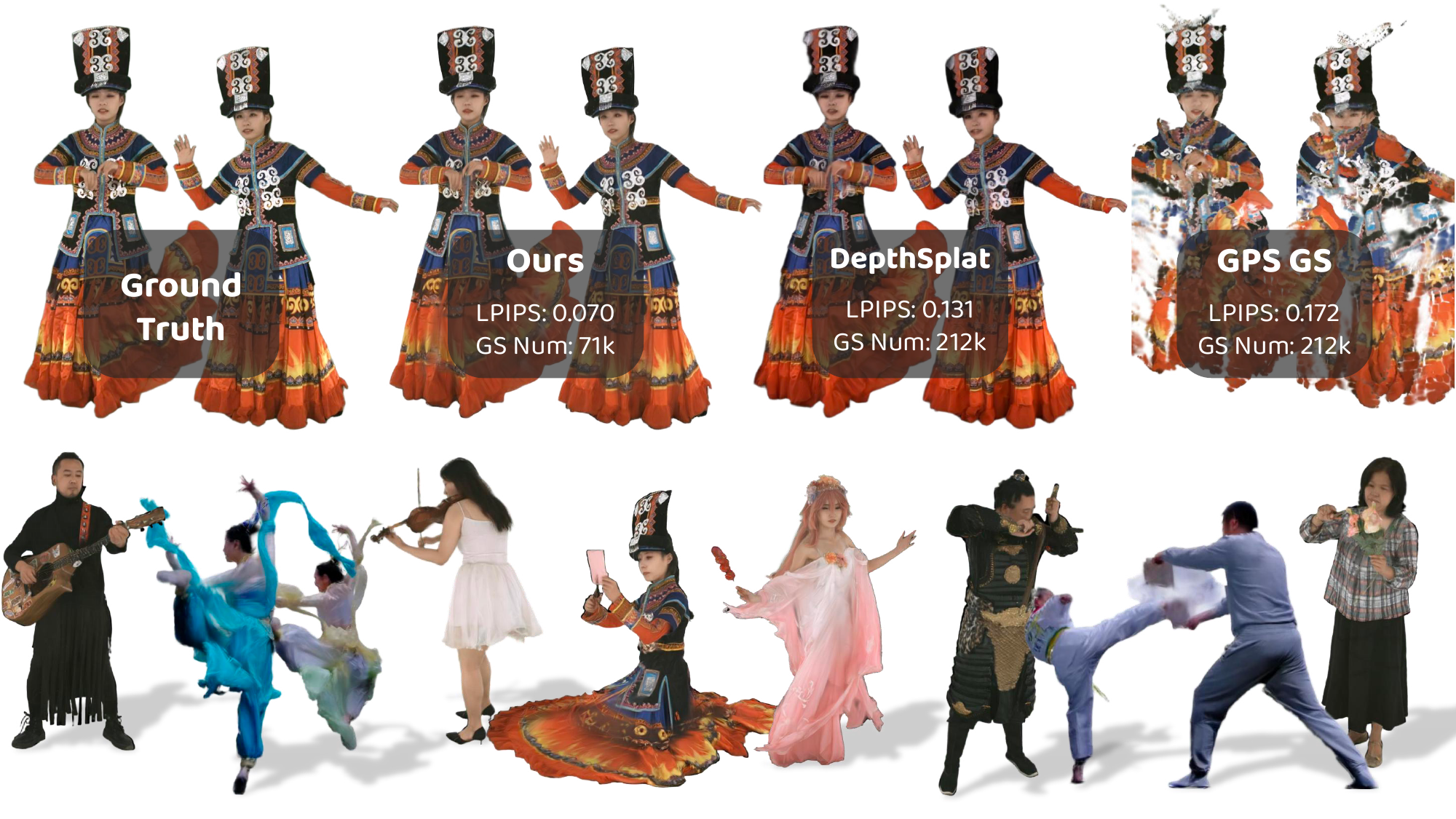}
\caption{Our method synthesizes high-fidelity novel views from sparse views by predicting a compact 3D Gaussian~\cite{kerbl3Dgaussians} representation.
The bottom row shows representative results of our method on diverse human performances.}
     \vspace{-0.1in}
    \label{fig:teaser}
\end{figure*}

Recently, neural reconstruction methods~\cite{kerbl3Dgaussians,zheng2024gpsgaussian,tang2024lgm,xu2024grm,gslrm2024,charatan23pixelsplat,chen2024mvsplat,xu2024depthsplat}
have demonstrated impressive view synthesis quality from sparse views,
significantly reducing the deployment cost of capture hardware.
These approaches typically adopt a \textit{view-centric} representation,
where pixel-aligned 3D features or Gaussian maps are predicted for each input view.
While this design enables fast reconstruction and real-time rendering from limited inputs,
it inherently introduces inter-view redundancy, where the same object is repeatedly represented across different views.
Such redundancy leads to rapidly increasing computation and transmission overhead as the number of views or rendering resolution grows,
limiting scalability under high-resolution and arbitrary-viewpoint settings.

\begin{figure}
    \centering
    \vspace{-0.15in}
    \includegraphics[width=0.8\linewidth]{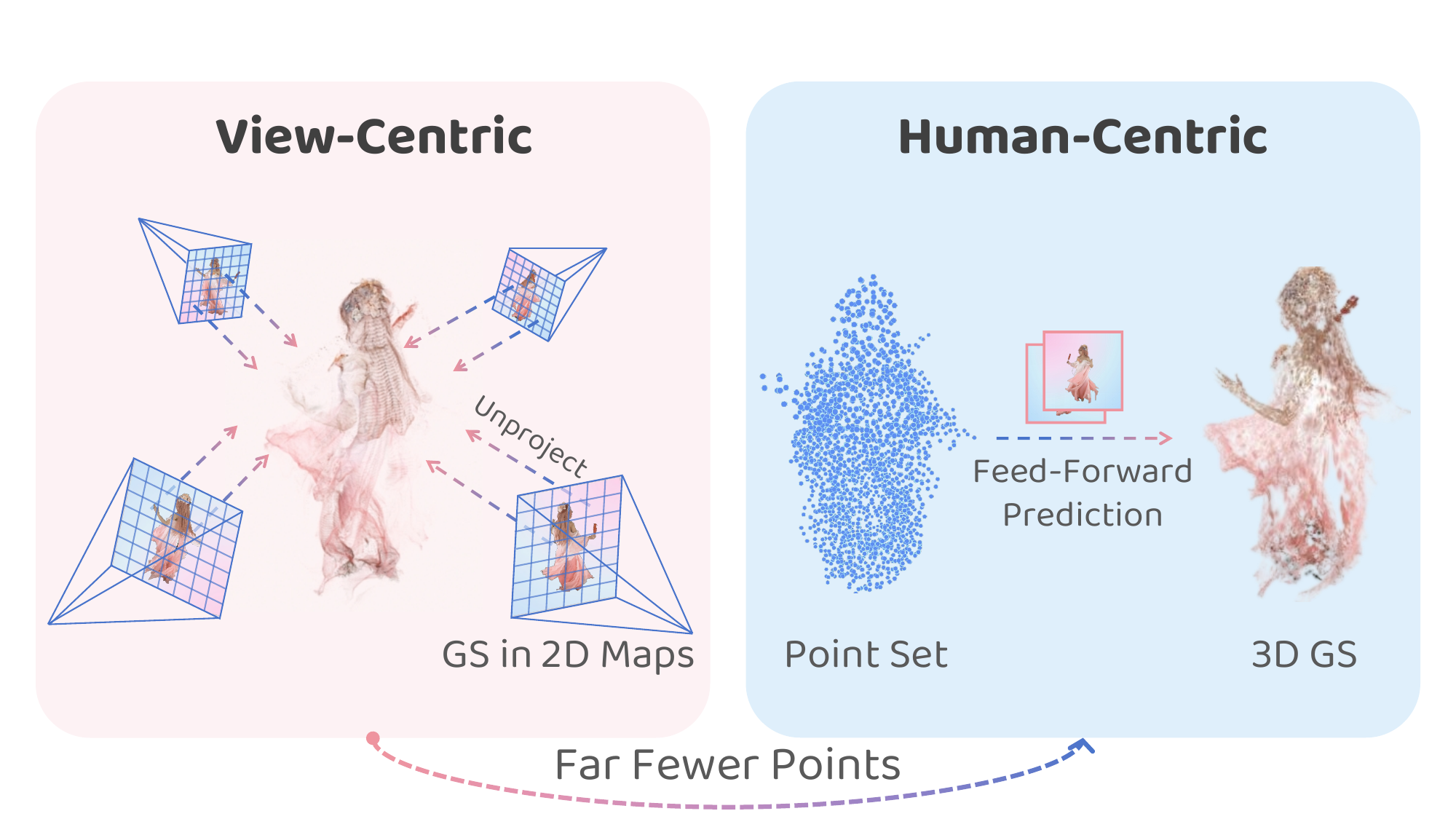}
    \vspace{-0.05in}
    \caption{\textbf{Comparison between view-centric and human-centric prediction.}
View-centric methods first predict pixel-aligned Gaussian maps for input views and then unproject them into 3D space.
In contrast, human-centric methods directly infer Gaussian primitives in 3D space by fusing multi-view observations.
    }
    \vspace{-0.1in}
    \label{fig:view_vs_3d}
\end{figure}

To overcome the redundancy inherent in view-centric representations,
we propose an \textit{human-centric} feed-forward approach that directly infers a compact Gaussian representation in 3D space, as illustrated in Figure~\ref{fig:view_vs_3d}.
This design not only reduces the number of Gaussians required but also improves the quality of novel view synthesis from sparse inputs.
Our pipeline first extracts appearance features from input images together with Plücker ray embeddings.
A coarse geometric proxy is then estimated via space carving~\cite{spacecarving1999} and voxelized to aggregate point-level features.
Finally, a Point-Image Transformer predicts the Gaussian parameters for each point in 3D space.
Benefiting from the geometric proxy, our method focuses prediction on 3D regions of interest,
fully exploiting the advantages of human-centric prediction,
which leads to substantially higher efficiency and improved robustness to variations in input view numbers and resolutions.


However, a remaining challenge is that the geometric hull often contains redundant interior points and lacks fine details, which degrades both prediction efficiency and quality.
To address this issue, we design a ray-casting mechanism to identify surface points within the geometric proxy.
This mechanism effectively removes internal points that do not contribute to rendering, thus improving prediction efficiency.
Moreover, by establishing explicit 2D–3D correspondences, it enriches point embeddings and enables more effective cross-modal interaction between image and point features, ultimately enhancing the quality of Gaussian prediction.

We conduct extensive experiments on various real-world (DNA-Rendering~\cite{2023dnarendering}, ActorsHQ~\cite{isik2023humanrf}, and PKU-DyMVHumans~\cite{zheng2024PKUDyMVHumans}) and synthetic (THuman2.0~\cite{tao2021function4d}, RenderPeople~\cite{renderpeople}) benchmarks, demonstrating that our method outperforms existing view-centric approaches in both efficiency and quality.
Notably, our approach achieves superior novel-view synthesis quality while using only about 33\% of the number of Gaussians required by view-centric methods.
Moreover, it exhibits strong robustness across varying input resolutions and view counts, highlighting its potential for practical applications.
In summary, our contributions are as follows:

\begin{enumerate}
    \item We introduce \sysname{}, a novel human-centric feed-forward approach that directly infers a compact Gaussian representation in 3D space, eliminating inter-view redundancy in view-centric methods.

    \item We propose a Point-Image Transformer that incorporates a ray-casting mechanism to remove redundant points and build explicit 2D–3D correspondences,
    unifying geometry and appearance modeling for efficient and accurate Gaussian prediction.

    \item We demonstrate that our approach consistently outperforms prior methods in both efficiency and reconstruction quality,
    while exhibiting strong robustness to variations in input view number and resolution.

\end{enumerate}

%% file: sec/2_related_work.tex
\section{Related Work}
\label{sec:related_work}

\PAR{Feed-forward 3D Reconstruction}
Feed-forward 3D reconstruction methods can be broadly categorized into implicit and explicit 3D representation learning approaches.
On one hand, implicit methods~\cite{yao2018mvsnet,wang2021ibrnet,lin2022enerf} leverage continuous functions that map input observations to target views.
Recently, LVSM~\cite{jin2025lvsm} learns to render novel views with minimal 3D bias,
but faces challenges in rendering efficiency, which is required for real-time applications.
On the other hand, explicit methods represent scenes with discrete geometric structures (e.g., point clouds), allowing direct geometric supervision and efficient rendering.
DUSt3R~\cite{dust3r_arxiv23} unifies monocular and stereo reconstruction through pairwise pointmap regression, offering flexibility but accumulating errors across multiple views.
\cite{wang2025vggt,wang2025pi3,depthanything3} extend this paradigm with unified transformer architectures that infer complete 3D attributes in a feed-forward manner.
However, their inherently view-centric design leads to inter-view redundancy and inconsistency in the final 3D representation.

\PAR{View-centric Representations}
View-centric methods typically reconstruct explicit geometry by predicting pixel-aligned representations for each input image.
Binocular methods~\cite{zheng2024gpsgaussian} unproject Gaussian maps from adjacent views into 3D Gaussians, while
MVS methods (e.g., Pixelsplat~\cite{charatan23pixelsplat}, MVSplat~\cite{chen2024mvsplat}, and DepthSplat~\cite{xu2024depthsplat})
use intermediate constraints and then unproject Gaussian maps from multiple views into a 3D representation.
LRM family~\cite{gslrm2024,xu2024grm,tang2024lgm} predicts per-view pixel-aligned 3D Gaussians from sparse images via transformers,
achieving fast feed-forward reconstruction in an end-to-end manner.
However, a common issue with these methods is that overlapping per-view predictions often cover the same regions multiple times,
leading to redundancy in the final representation.
Post-processing strategies like Gaussian pruning~\cite{ziwen2025llrm,zhang2024GGN} or feed-forward compression~\cite{chen2025fcgs}
are proposed to reduce redundancy but introduce ambiguities in predicting dominant or low-opacity results on specific views.
Similarly, generative approaches~\cite{szymanowicz2025bolt3d,gao2024cat3d} generate Gaussian maps or multi-view images before lifting them into 3D,
producing diverse results but still suffering from redundancy and ambiguities due to the view-centric design.

\PAR{Native 3D Representations}
Several distinct approaches have been developed for 3D reconstruction, each with unique strengths and limitations.
Radiance field-based methods~\cite{xu2022point} build on the implicit representation of 3D geometry using dense input point clouds.
However, they are computationally expensive and depend heavily on dense, structured data.
KPlanes-based methods~\cite{instant3d2023,hong2023lrm} use multi-plane features for efficient 3D reconstruction, yet learning consistent view-to-plane mappings remains challenging.
For human-centric applications, approaches like LHM and LHM++~\cite{qiu2025LHM,qiu2025lhmpp} leverage a parametric model as a shape prior, enabling the prediction of 3D Gaussians.
While this offers high precision for human models, its reliance on SMPL restricts its applicability to complicated scenes.

Voxel-based optimization methods~\cite{yu2022plenoxels,lu2024scaffoldgs,ren2024octree} further explore explicit 3D representations by discretizing scenes into voxel grids.
Recently, AnySplat~\cite{jiang2025anysplat} employs differentiable voxelization to aggregate 3D Gaussians in a feed-forward manner.
Although effective in reducing redundancy, such designs rely on additional geometric supervision,
which may limit their adaptability to diverse scenarios.

Beyond the reconstruction task, recent advances in 3D generation~\cite{zhang20233dshape2vecset,hunyuan3d22025tencent,wu2024direct3d,zhang2024clay}
have explored learning object-centric representations directly from data.
These methods typically employ VAEs to encode point clouds into a latent space for generating Signed Distance Functions (SDFs).
While such generative frameworks excel at producing coherent 3D structures,
they are not designed for multi-view reconstruction and struggle to ensure consistency with 2D observations.
Nonetheless, they highlight the potential of compact and unified 3D representations—a goal our approach aims to achieve in a reconstruction context.

%% file: sec/3_method.tex
\section{Method}

\begin{figure*}[ht!]
    \centering
    \vspace{-0.2in}
    \includegraphics[width=0.99\linewidth]{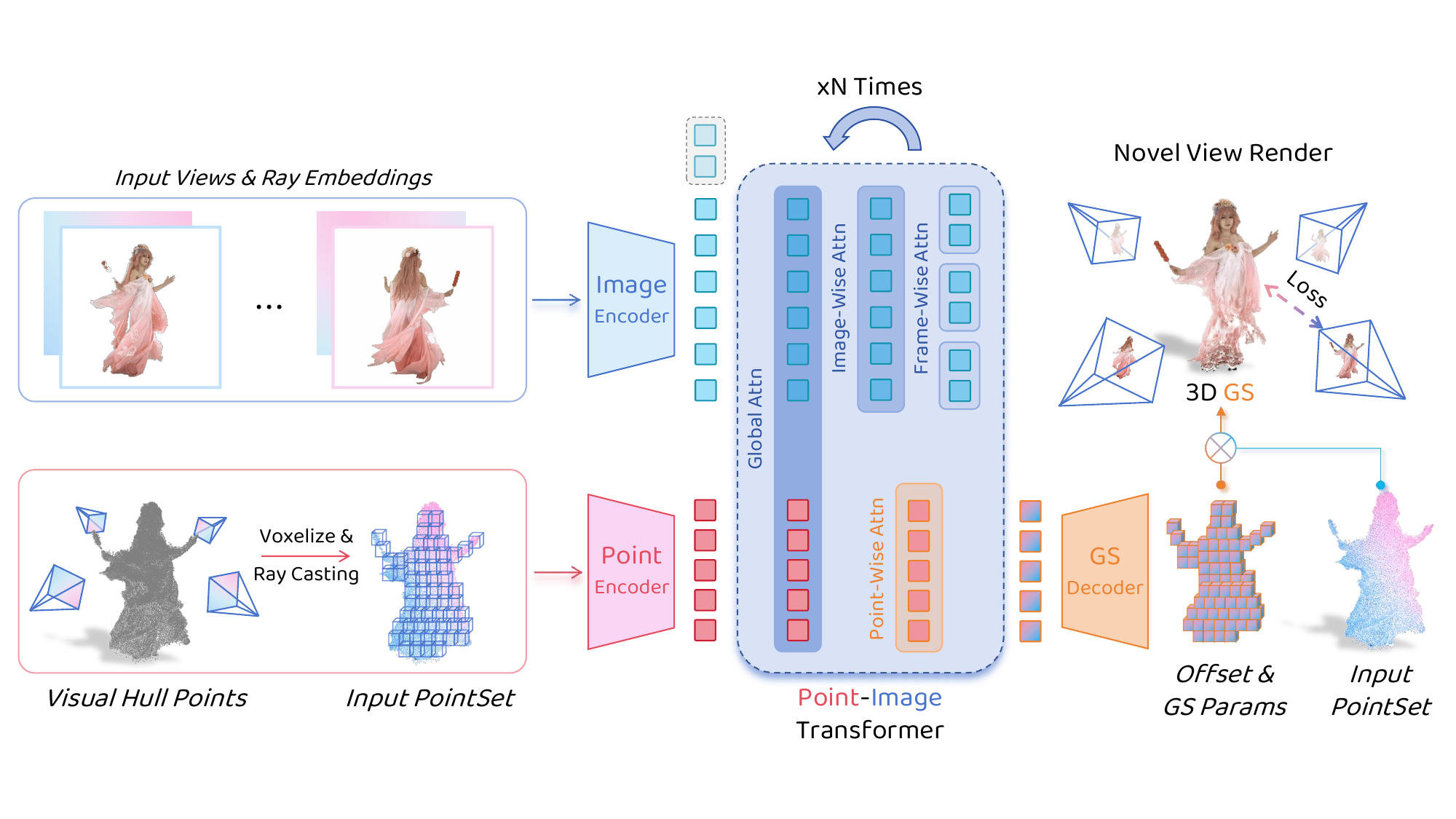}
    \vspace{-0.2in}
    \caption{\textbf{Pipeline Overview.}
    Given calibrated multi-view images and masks, \sysname{} reconstructs a compact 3D Gaussian (3DGS) representation in a feed-forward step.
    We first extract appearance features from input images together with Plücker ray embeddings.
    A coarse geometric proxy is then estimated via space carving, followed by voxelization for structured representation.
    For each anchor view, we perform ray casting to select surface points and build explicit 2D–3D correspondences.
    A Point-Image Transformer then integrates appearance and geometry features and predicts point offsets and the remaining Gaussian parameters,
    constructing the final 3DGS representation.
    The entire framework is trained end-to-end under RGB supervision only, enabling efficient reconstruction.
    }
    \label{fig:pipeline}
\end{figure*}

Given a set of calibrated images and corresponding object masks, our method aims to reconstruct a compact 3D Gaussian Splatting (3DGS) representation in a feed-forward manner.
The overview of the proposed method is illustrated in Figure~\ref{fig:pipeline}.
It begins with the extraction of appearance features from 2D images and camera rays (Section \ref{sec:appearance_extraction}).
Next, a coarse geometric proxy is estimated to ensure visibility consistency.
To reduce redundancy and establish 2D--3D correspondence, we employ ray casting with ray embeddings, which unify 2D and 3D representations (Section \ref{sec:geometry_extraction}).
Subsequently, a Point-Image Transformer fuses appearance and geometry features and decodes them into Gaussian parameters (Section \ref{sec:point_image_fusion}).
The entire network is trained end-to-end, requiring only RGB supervision (Section \ref{sec:end_to_end_training}).

\subsection{Appearance Feature Extraction}
\label{sec:appearance_extraction}

We extract 3D-aware appearance features from 2D images by embedding Plücker rays into the appearance representation.

\PAR{Plücker Ray Embedding}
Each pixel is represented by its Plücker ray, encoding the ray direction and position. The Plücker coordinates are defined as:
\begin{equation}
\mathbf{f}_{\mathrm{ray}} = [\mathbf{d}, \mathbf{o} \times \mathbf{d}] \in \mathbb{R}^6,
\end{equation}
where $\mathbf{o}$ and $\mathbf{d}$ are the camera origin and ray direction, respectively. These are concatenated with the pixel color $\mathbf{c}$ to form:
\begin{equation}
\mathbf{f}_{\mathrm{pixel}} = [\mathbf{c}, \mathbf{f}_{\mathrm{ray}}] \in \mathbb{R}^9.
\end{equation}
Images are patchified into a token sequence:
\begin{equation}
\mathbf{t}_{\mathrm{images}} = \text{Patchify}(\mathbf{f}_{\mathrm{pixel}}) \in \mathbb{R}^{N \times C}.
\end{equation}

\PAR{Masked Token Sampling}
We apply mask-guided top-$n$ sampling to select informative tokens.
The score for each token is computed by summing the mask values within its patch:
\begin{equation}
s_i = \sum_{j \in \text{Patchify}(i)} m_j.
\end{equation}
The top-$n$ tokens are then selected based on these scores:
\begin{equation}
\mathbf{t}_{\mathrm{appearance}} = \text{SelectTop-}n(\{t_i\}, \{s_i\}).
\end{equation}
Appearance tokens are then projected into a hidden dimension $C$ using a linear layer:
\begin{equation}
\mathbf{T}_{\mathrm{appearance}} = \text{Linear}(\mathbf{t}_{\mathrm{appearance}}) \in \mathbb{R}^{n \times C}.
\end{equation}

\subsection{Geometry Feature Extraction}
\label{sec:geometry_extraction}

We aim to obtain reliable 3D geometry that can serve as a spatial reference for appearance features.
Unlike methods that rely solely on depth estimation, which often suffer from noise and inconsistency, our approach combines space carving and ray casting to build accurate 2D–3D correspondences.

\PAR{Geometry Proxy}
We first construct a visibility-consistent proxy of the object using a space carving algorithm~\cite{spacecarving1999}.
Given the calibrated masks from multiple views, we iteratively remove points that are not visible in all masks, producing a dense visual hull that approximates the object surface.
The resulting point cloud is denoted as $\mathbf{p}$ and serves as the geometric basis for subsequent voxelization.

\PAR{Voxelization and Ray Casting}
To remove redundant interior points and establish explicit 2D–3D correspondences, we perform ray casting from each anchor view.
A voxel grid is first constructed to enclose the input point cloud $\mathbf{p}$, forming a structured spatial partition of the scene.
We denote the resulting voxelized points as $\mathbf{p}_v$.
For each pixel ray, parameterized as $\mathbf{r}(t) = \mathbf{c} + t\mathbf{d}$, with $\mathbf{c}$ as the camera center and $\mathbf{d}$ as the ray direction, we apply a Digital Differential Analyzer (DDA) to efficiently traverse the grid.
During traversal, for each voxel intersected by the ray, we identify the point within that voxel that lies closest to the ray by minimizing the point-to-ray distance:
\begin{equation}
d(\mathbf{p}_v, \mathbf{r}) = \frac{\|(\mathbf{p}_v - \mathbf{c}) \times \mathbf{d}\|}{\|\mathbf{d}\|}, \quad
\mathbf{p}_v^* = \arg\min_{\mathbf{p}_v \in v} d(\mathbf{p}_v, \mathbf{r}).
\end{equation}
Once the ray intersects a voxel, we select a fixed number of points $s$ from the voxel.
Voxels not intersected by any ray are discarded.
We denote the resulting set of points as $\mathbf{p}_s^*$.
This facilitates explicit image-to-point correspondences while removing redundant internal geometry.

\PAR{Point Encoding}
Each voxel, which contains a subset of 3D points, is first hashed to identify the occupied regions in the voxel grid.
For every occupied voxel, we apply a sinusoidal positional encoding to the points it contains.
The encoded point features $\gamma(\mathbf{p}_s^*)$ are then concatenated with the corresponding ray feature $\mathbf{f}_{\mathrm{ray}} \in \mathbb{R}^6$ cast from the 2D anchor view.
This representation is further projected through a linear layer followed by layer normalization~\cite{ba2016layernormalization} to produce the geometry token:
\begin{equation}
\mathbf{T}_{\mathrm{geometry}} = \text{LN}\left(\text{Linear}\big([\gamma(\mathbf{p}_s^*); \mathbf{f}_{\mathrm{ray}}]\big)\right) \in \mathbb{R}^{M \times C}.
\end{equation}
Here, $\gamma: \mathbb{R}^3 \to \mathbb{R}^{3L}$ applies an $L$-frequency sinusoidal encoding to spatial coordinates, and $M$ denotes the number of occupied voxels.
Through this process, we obtain a unified embedding that jointly represents appearance and geometry, facilitating effective cross-modal modeling.

\subsection{3DGS Prediction}
\label{sec:point_image_fusion}
\PAR{Point-Image Transformer}
The fusion of 2D and 3D information is achieved through a transformer tailored for point-image interaction.
It takes as input the appearance tokens $\mathbf{T}_{\mathrm{appearance}}$ and the geometric tokens $\mathbf{T}_{\mathrm{geometry}}$,
and outputs the fused Gaussian Splatting tokens $\mathbf{T}_{\mathrm{Gaussians}}$:
\begin{equation}
\mathbf{T}_{\mathrm{Gaussians}} = \text{Network}(\mathbf{T}_{\mathrm{appearance}}, \mathbf{T}_{\mathrm{geometry}}),
\end{equation}
where the network employs alternating attention~\cite{wang2025vggt} to integrate global, point-wise, and image-wise information.
\begin{itemize}
    \item \textbf{Global Attention} Self-attention is performed to capture holistic context by enabling interactions among all tokens (appearance and geometry).
    \item \textbf{Point-wise Attention} We then use full self-attention to enhance local geometric coherence by focusing on neighboring point sets.
    \item \textbf{Image-wise Attention} Self-attention is further used to model intra-view relationships (frame-wise attention) and inter-view dependencies (cross-view attention) for multi-view aggregation.
\end{itemize}
The resulting $\mathbf{T}_{\mathrm{Gaussians}}$ serves as the input for the 3DGS parameter prediction module.

\PAR{3DGS Parameter Prediction}
The fused token features are decoded into Gaussian splatting parameters, including position offset, scale, rotation, spherical harmonics (SH) coefficients, and opacity.

The decoding process is formulated as:
\begin{equation}
\mathbf{T}_{\mathrm{Gaussians}}: \{g_i\}_{i=1}^M
\;\;\rightarrow\;\;
\left\{
  \left\{
    \big(o_i^k,\, c_i^k,\, s_i^k,\, \alpha_i^k,\, r_i^k\big)
  \right\}_{k=1}^K
\right\}_{i=1}^M,
\end{equation}
where each $g_i$ is decoded into $K$ Gaussians with
position offsets $o$, colors $c$, scales $s$, opacities $\alpha$, and rotations $r$.

Each parameter is predicted using a separate linear layer with different activation functions to ensure valid ranges.
For each property, we use a variable $v \in \{o, c, s, \alpha, r\}$. The prediction can be written as:
\begin{equation}
v = f_v\big(\text{Linear}_v(\mathbf{T}_{\mathrm{Gaussians}})\big) \in \mathbb{R}^{K \times D_v}
\end{equation}
where $f_v$ is the activation function for property $v$, and $D_v$ is the dimension of property $v$.
The predicted offsets are added to the input point set positions to obtain the final Gaussian centers.
\begin{equation}
\mathbf{p}_{\mathrm{final}} = \mathbf{p}_{s}^* + o
\end{equation}

\subsection{Training}
\label{sec:end_to_end_training}

We train the entire framework end-to-end using only RGB supervision, enabling joint refinement of geometry and appearance without additional priors.
The overall objective combines pixel-wise and perceptual losses:
\begin{equation}
\mathcal{L} = \lambda_{\mathrm{L1}} \mathcal{L}_{\mathrm{L1}} + \lambda_{\mathrm{LPIPS}} \mathcal{L}_{\mathrm{LPIPS}},
\end{equation}
where $\mathcal{L}_{\mathrm{L1}}$ enforces photometric consistency and $\mathcal{L}_{\mathrm{LPIPS}}$ promotes perceptual similarity.
$\lambda_{\mathrm{L1}}$ and $\lambda_{\mathrm{LPIPS}}$ balance the two terms.



%% file: sec/4_experiments.tex
\section{Experiments}
\label{sec:experiments}

\subsection{Implementation Details}
We normalize the scene to a $2\times2\times2$ cube.
The appearance branch operates on $4\times4$ image patches, while the geometry branch uses voxels of size $0.005$.
We use a Point-Image Transformer with 4 blocks and a hidden size of 1024.
During decoding, we set the number of per-token Gaussians $K$ to $16$.
QK-Norm~\cite{henry2020qknorm} is applied to every attention layer for stability, following prior work~\cite{gslrm2024,jin2025lvsm}.
The initial learning rate is $2\times10^{-4}$ with a linear warmup over the first 10\% of iterations.
$\lambda_{\mathrm{L1}}$ and $\lambda_{\mathrm{LPIPS}}$ are set to $1.0$ and $1.0$, respectively.
Training efficiency is improved using FlashAttention~\cite{dao2023flashattention2}, gradient checkpointing, and mixed precision.
We train for 300k iterations with a total batch size of 32 on NVIDIA H20 GPUs. All inference timing is measured on an NVIDIA A6000 (48GB).

\subsection{Datasets and Metrics}

We train our model on DNA-Rendering~\cite{2023dnarendering}, using approximately $1{,}000$ sequences for training and $16$ sequences for evaluation, following prior work~\cite{jin2025diffuman4d}, unless otherwise specified. To assess performance in multi-human scenarios, we evaluate the proposed method using the PKU-DyMVHumans~\cite{zheng2024PKUDyMVHumans} dataset. Furthermore, we investigate the zero-shot generalization capability of the model on the ActorsHQ~\cite{isik2023humanrf} dataset, which consists of $12$ dynamic human sequences.
Following prior work on human novel-view synthesis~\cite{peng2021neural,lin2022enerf,zheng2024gpsgaussian}, we report PSNR, SSIM~\cite{wang20024ssim}, and LPIPS~\cite{zhang2018perceptual} computed on foreground regions defined by the subject's bounding box.
For efficiency, we also report the number of Gaussians and the per-frame inference time.

\subsection{Comparison with State-of-the-art Methods}
\PAR{Baselines.}
We compare the proposed approach against state-of-the-art feed-forward NVS methods. These include view-centric methods, namely GS-LRM~\cite{gslrm2024}, DepthSplat~\cite{xu2024depthsplat}, LVSM~\cite{jin2025lvsm}, and GPS-Gaussian~\cite{zheng2024gpsgaussian}, the pose-free baseline AnySplat~\cite{jiang2025anysplat}, as well as human-specific approaches, such as RoGSplat~\cite{RoGSplat2025CVPR} and LHM~\cite{qiu2025LHM}. For DNA-Rendering comparisons, all trainable baselines are trained or fine-tuned on DNA-Rendering with a matched optimization budget. For Table~\ref{tab:comparison_synthetic_dataset}, we train a separate model on THuman2.0 with the same training-set size, since GPS-Gaussian and RoGSplat are trained on THuman2.0 and require ground-truth depth supervision. Furthermore, we compare our approach with optimization-based techniques, specifically 4DGS~\cite{yang2023gs4d,xu2024longvolcap} and GauHuman~\cite{GauHuman}, along with the state-of-the-art generative frameworks LGM~\cite{tang2024lgm} and Diffuman4D~\cite{jin2025diffuman4d}.

\begin{figure*}[ht!]
    \vspace{-2mm}
    \centering
    \includegraphics[width=0.99\linewidth]{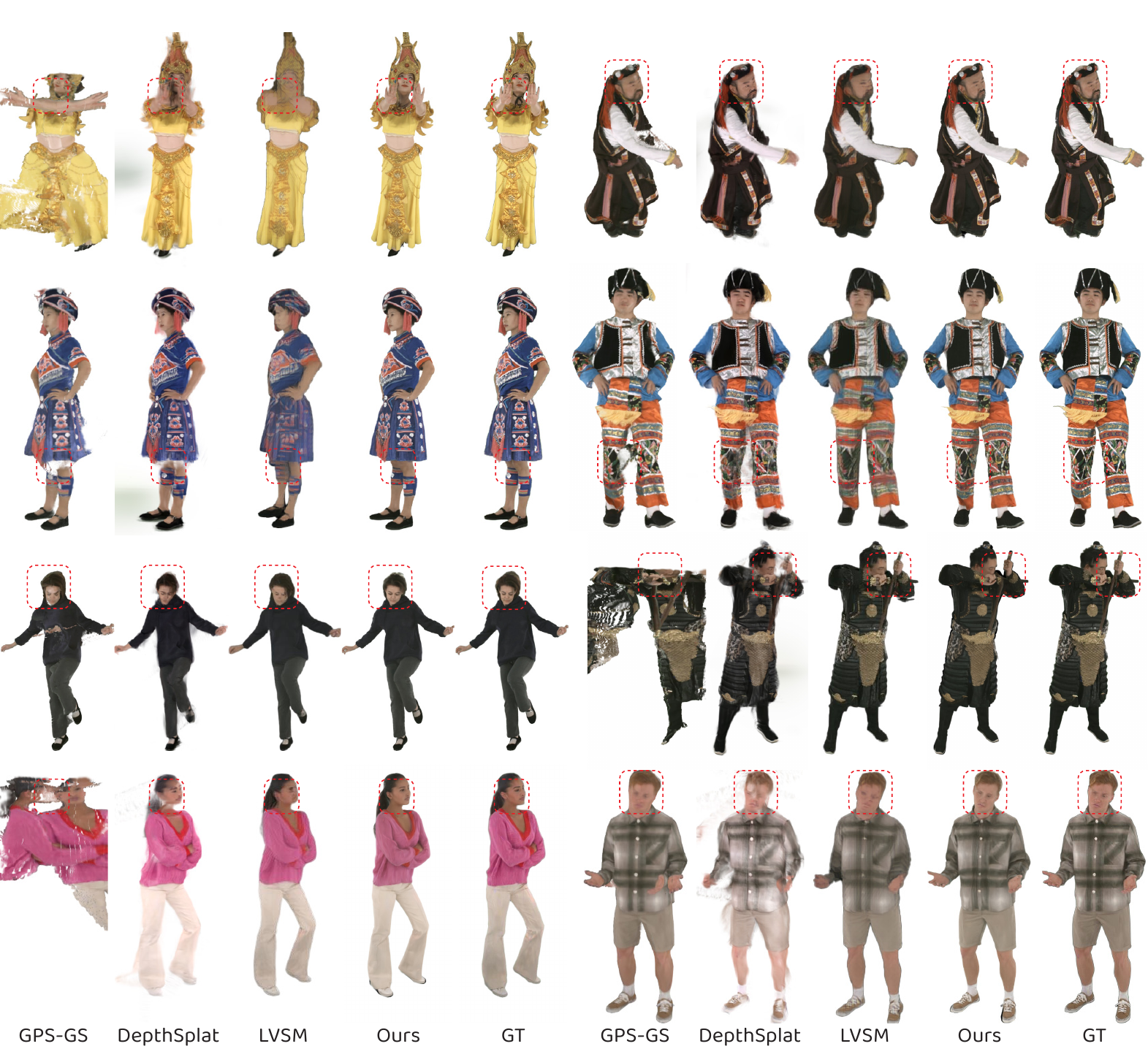}
    \caption{\textbf{Qualitative comparison across datasets}.
Our method produces compact and complete 3D reconstructions that yield clean and consistent novel-view renderings.
View-centric baselines (e.g., GPS-Gaussian, DepthSplat) show redundant or missing geometry, resulting in floating artifacts,
while LVSM exhibits over-smoothed and mosaic-like appearances.}
    \label{fig:comparison}
\end{figure*}

\begin{figure}
    \centering
    \includegraphics[width=0.8\linewidth]{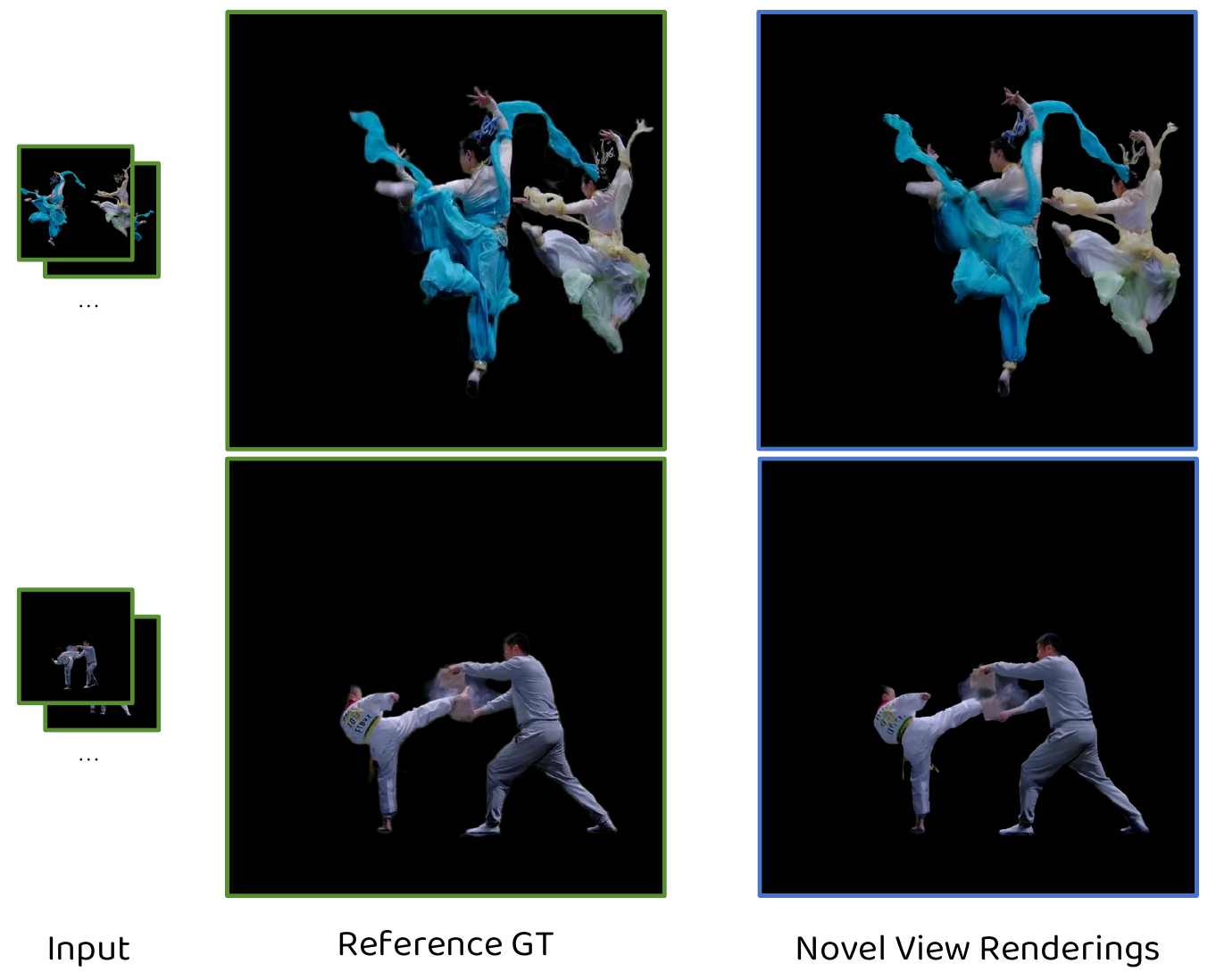}
    \vspace{-0.1in}
    \caption{\textbf{Qualitative examples of novel view synthesis on the DyMVHumans dataset.}
Our method generates high-quality novel views for multi-human scenes with diverse poses and appearances using eight input views.}
    \vspace{-1mm}
    \label{fig:multicentric}
\end{figure}

\input{tables/comparison_dataset}
\PAR{Results on different datasets.}
We report results on the DNA-Rendering and ActorsHQ datasets in Table~\ref{tab:comparison_dataset} and Fig.~\ref{fig:comparison}.
All methods are evaluated at 512$\times$512 resolution using 8 input views, with pure PyTorch implementations (without custom CUDA kernels for acceleration).
As shown by both quantitative metrics and qualitative comparisons, our method achieves the best overall performance across both datasets.
Our approach directly infers compact 3D Gaussians, enabling consistent novel view synthesis without redundant structures or unmodeled regions.
In contrast, GS-LRM, GPS-Gaussian, and DepthSplat exhibit redundancy across views, leading to floating artifacts and noisy points when rendering novel views.
LVSM can model continuous novel views but tends to produce over-smoothed, mosaic-like results and suffers from slow rendering.
AnySplat operates without pose priors, which partly explains its lower fidelity on this human-centric benchmark.
We further demonstrate our method's effectiveness on multi-human-centric dynamic scenes from the PKU-DyMVHumans dataset in Fig.~\ref{fig:multicentric}.

\PAR{Results on high resolution.}
We further evaluate at 1024$\times$1024 on the DNA-Rendering test set in a zero-shot setting, without additional fine-tuning.
For the 4DGS baseline~\cite{yang2023gs4d}, we adopt the reconstruction and rendering framework described in~\cite{xu2024longvolcap} and synthesize novel views for evaluation.
As shown in Table \ref{tab:comparison_1024}, our method maintains clear superiority over other feed-forward methods at higher resolution,
demonstrating robust zero-shot scaling to detailed human performances.
Our method achieves comparable PSNR, SSIM, and LPIPS while being much faster compared to Diffuman4D.
Moreover, even when compared with state-of-the-art dense-view (e.g., 44 views) optimization-based 4DGS methods,
it delivers similar visual fidelity with a fraction of the computational cost.
Note that LVSM cannot handle this resolution due to memory constraints.
\input{tables/comparison_1024}

\PAR{Results on different view numbers.}
We further evaluate the zero-shot generalization ability by varying the number of input views.
As shown in Table \ref{tab:comparison_camnum}, our method consistently outperforms all baselines under different view counts,
demonstrating strong robustness to view number variation, which can be attributed to its object-centric design.

\input{tables/comparison_camnum}

\subsection{Ablation Study}

\begin{figure}
    \centering
    \includegraphics[width=0.8\linewidth]{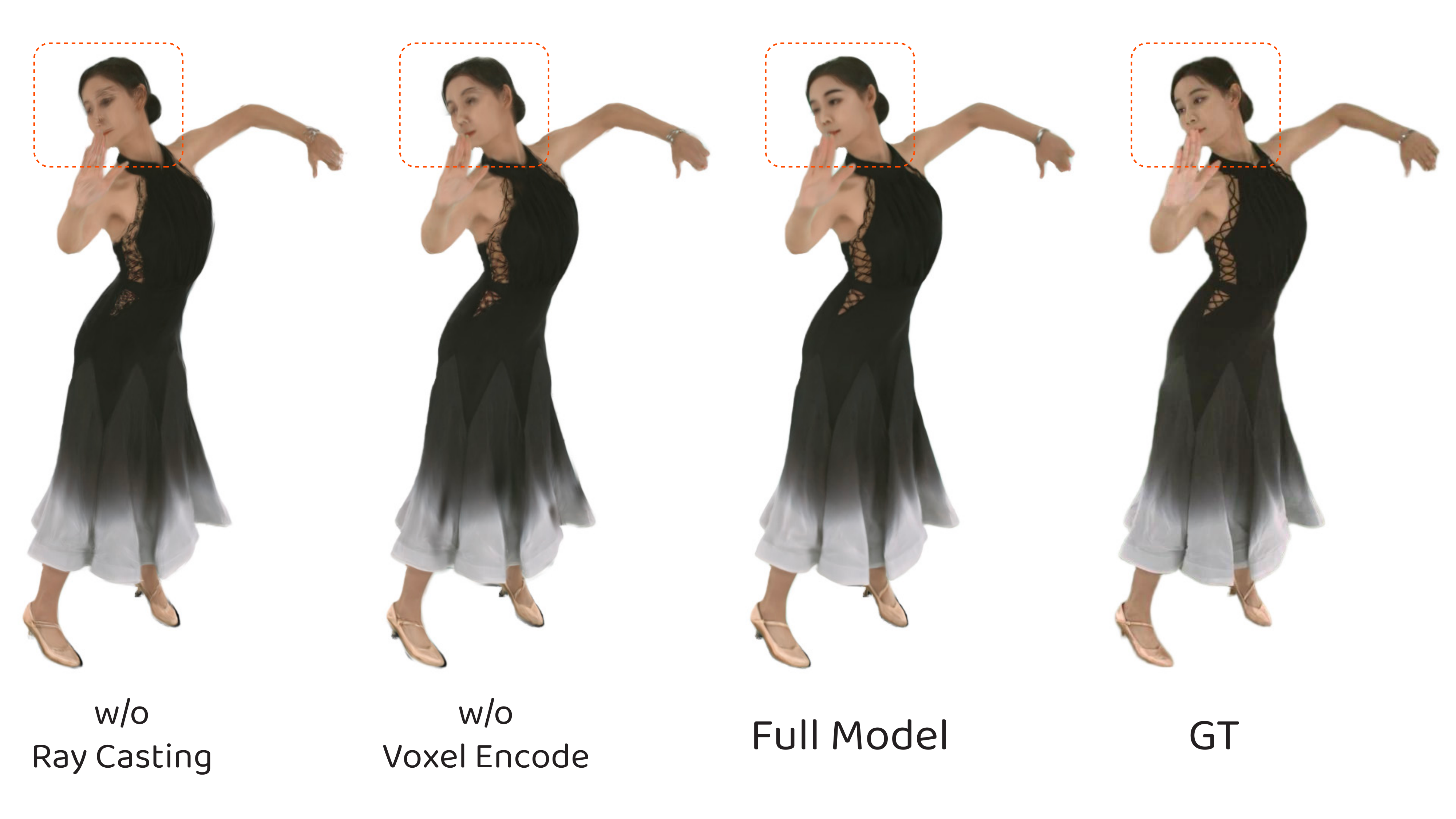}
    \vspace{-0.1in}
    \caption{\textbf{Ablation study of modules.}
    Removing our proposed components leads to visible artifacts in the rendered results, especially in challenging regions such as the face.
    }
    \vspace{-0.1in}
    \label{fig:ablation_modules}
\end{figure}

\PAR{Effects of Modules.}
We conduct ablation studies to investigate the contribution of each component in our approach, as reported in Table~\ref{tab:ablation_modules} and Table~\ref{tab:ablation_arch}.
Each variant is obtained by selectively removing or altering a specific module to assess its individual effect.
As illustrated in Fig.~\ref{fig:ablation_modules}, removing the point-encoding module leads to noticeable artifacts and inconsistencies in the rendered views,
as the model struggles to learn a unified 3D representation from multiple views.
Omitting the ray-casting mechanism results in degradation of high-frequency details and introduces internal redundancy.
Finally, as shown in Table~\ref{tab:ablation_arch},
the proposed Point-Image Transformer attains better quality with full self-attention transformers~\cite{jin2025lvsm,gslrm2024}, yet with markedly lower computational overhead.
\input{tables/ablation_modules}

\PAR{Geometry Proxy Type}
We investigate the impact of different geometry proxy types on the performance of our method.
The results are summarized in Table \ref{tab:ablation_geometry}, where we compare the performance using various geometry proxy types.
VGGT Depth indicates using depth maps predicted by a pre-trained VGGT~\cite{wang2025vggt} model as the geometry proxy.
Bounding Box indicates using the foreground bounding box as the geometry proxy.
Frustum Sample indicates sampling 3D points along camera rays within a predefined frustum as the geometry proxy.
Our findings indicate that the consistent geometry proxy (e.g., Visual Hull) enhances the rendering quality and robustness.

\PAR{Voxel Size}
We explore the influence of voxel size on the performance of our method.
The results are presented in Table \ref{tab:ablation_voxel_size}, where we evaluate different voxel sizes and their corresponding effects on rendering quality.
In our experiments, we choose a voxel size of $0.0050$, which achieves the best trade-off between performance and efficiency.

\input{tables/ablation_arch_geometry}
\input{tables/ablation_voxel_size}

%% file: tables/comparison_dataset.tex
\begin{table*}[!t]
\centering
\FloatBarrier
\renewcommand{\arraystretch}{1.2}
\caption{\textbf{Quantitative comparison on the DNA-Rendering and ActorsHQ datasets.}
$\times N$ denotes that the metric scales linearly with $N$ target views.
“GS-Num” represents the number of Gaussians, and “Time” refers to inference time (in seconds).
All methods are evaluated with eight $512 \times 512$ input views.
The best results are highlighted in bold.
}
\begin{tabular}{lcccccccc}
\toprule
& \multicolumn{3}{c}{\textbf{DNA-Rendering}~\cite{2023dnarendering}}
& \multicolumn{3}{c}{\textbf{ActorsHQ}~\cite{isik2023humanrf}}
& \multirow{2}{*}{\textbf{GS-Num}}
& \multirow{2}{*}{\textbf{Time}} \\
\cmidrule(lr){2-7}
\textbf{Method} & PSNR$\uparrow$ & SSIM$\uparrow$ & LPIPS$\downarrow$
& PSNR$\uparrow$ & SSIM$\uparrow$ & LPIPS$\downarrow$ & \\
\midrule
GS-LRM~\cite{gslrm2024}                       & 18.25	& 0.582	& 0.364 & 16.96	& 0.675	& 0.299 & 204k           & \textbf{0.32} \\
AnySplat~\cite{jiang2025anysplat}                & 20.97 & 0.790 & 0.143 & 19.11 & 0.699 & 0.238 & 86k & 0.53 \\
GPS-Gaussian~\cite{zheng2024gpsgaussian}                & 22.35 & 0.797 & 0.172 & 23.81 & 0.887 & 0.084 & 51k $\times$ N & 0.07 $\times$ N \\
DepthSplat~\cite{xu2024depthsplat}                    & 23.98	& 0.821	& 0.131 & 22.39	& 0.797	& 0.157 & 212k           & 0.45 \\
LVSM~\cite{jin2025lvsm}                         & 23.24	& 0.802	& 0.135 & 24.00	& 0.804	& 0.128 & -              & 1.0 $\times$ N  \\
Ours                        & \textbf{27.18} & \textbf{0.891} & \textbf{0.071} & \textbf{27.62} & \textbf{0.887} & \textbf{0.084} & \textbf{71k} & 0.40 \\
\bottomrule
\end{tabular}

\label{tab:comparison_dataset}
\end{table*}

\begin{table*}[!t]
\centering
\FloatBarrier
\renewcommand{\arraystretch}{1.2}
\caption{\textbf{Quantitative comparison on synthetic datasets.}
The evaluation metrics are computed across the entire images, following the protocol established in~\cite{RoGSplat2025CVPR}.
}
\begin{tabular}{lccccccccc}
\toprule
& \multirow{2}{*}{\textbf{View}}
& \multicolumn{3}{c}{\textbf{THuman2.0}~\cite{tao2021function4d}}
& \multicolumn{3}{c}{\textbf{RenderPeople}~\cite{renderpeople}}
& \multirow{2}{*}{\textbf{GS-Num}} \\
\cmidrule(lr){3-8}
\textbf{Method} &  & PSNR$\uparrow$ & SSIM$\uparrow$ & LPIPS$\downarrow$
& PSNR$\uparrow$ & SSIM$\uparrow$ & LPIPS$\downarrow$ \\
\midrule
RoGSplat~\cite{RoGSplat2025CVPR}     & 4       & 28.94 & 0.961 & 0.043 & 25.12 & 0.938 & 0.066 & 135k \\
GPS-Gaussian        & 6       & 27.79 & 0.960 & 0.039 & 25.11 & 0.932 & 0.068 & 67k $\times$ N \\
Ours                & 4       & \textbf{30.68} & \textbf{0.969} & \textbf{0.021} & \textbf{26.70} & \textbf{0.954} & \textbf{0.051} & \textbf{89k}\\
\midrule
DepthSplat          & 8       & 26.32 & 0.951 & 0.048 & 26.90 & 0.958 & 0.051 & 270k \\
GPS-Gaussian        & 8       & 31.08  & 0.970 & 0.032 & 29.97 & 0.970 & 0.035 & 67k $\times$ N \\
Ours                & 8       & \textbf{34.30} & \textbf{0.980} & \textbf{0.015} & \textbf{32.36} & \textbf{0.974} & \textbf{0.026} & \textbf{75k} \\
\bottomrule
\end{tabular}

\vspace{-2mm}
\label{tab:comparison_synthetic_dataset}
\end{table*}


%% file: tables/comparison_1024.tex
\begin{table}
\small
\caption{\textbf{Quantitative comparison at 1024 resolution on the DNA-Rendering test set.}
"opt" indicates that the method is optimization-based, whereas "prior" denotes the utilization of templates such as SMPL.
}
\vspace{-1mm}
\centering
\setlength{\tabcolsep}{2.3pt}
\begin{tabular}{l|c|c|ccc}
\toprule

\multicolumn{1}{c}{Method} & \multicolumn{1}{c}{Type} & \multicolumn{1}{c}{View} & PSNR$\uparrow$ & SSIM$\uparrow$ & LPIPS$\downarrow$ \\

\midrule

LongVolCap(4DGS)~\cite{xu2024longvolcap}         & opt  & 8      & 24.21 & 0.840 & 0.221 \\

GauHuman~\cite{GauHuman} & prior, opt & 8 & 17.60 & 0.714 & 0.270 \\

Diffuman4D~\cite{jin2025diffuman4d}   & prior, opt  & 8      & 26.32 & 0.881 & 0.150 \\

\midrule

LHM~\cite{qiu2025LHM} & prior, feed-forward & 1 & 19.73 & 0.751 & 0.242 \\

LGM~\cite{tang2024lgm} & feed-forward & 4 & 17.18 & 0.698 & 0.311 \\
GPS-GS~\cite{zheng2024gpsgaussian} & feed-forward  & 8      & 20.63 & 0.779 & 0.225 \\
DepthSplat~\cite{xu2024depthsplat} & feed-forward  & 8      & 19.82 & 0.704 & 0.328 \\

Ours & feed-forward & 8      & \textbf{26.54} & \textbf{0.882} & \textbf{0.123} \\

\bottomrule
\end{tabular}


\label{tab:comparison_1024}
\end{table}

%% file: tables/comparison_camnum.tex
\begin{table}
\small
\caption{\textbf{Quantitative comparison with different numbers of input cameras.}
We evaluate the zero-shot generalization of our method to different numbers of input views.
Our method remains robust when the number of input views changes.
}
\vspace{-1mm}
\centering
\setlength{\tabcolsep}{1pt}
\begin{tabular}{lcccccc}
\toprule
& \multicolumn{3}{c}{4 cameras}
& \multicolumn{3}{c}{16 cameras} \\
\cmidrule(lr){2-7}
\textbf{Method} & PSNR$\uparrow$ & SSIM$\uparrow$ & LPIPS$\downarrow$ & PSNR$\uparrow$ & SSIM$\uparrow$ & LPIPS$\downarrow$ \\
\midrule
GPS-GS~\cite{zheng2024gpsgaussian}  & 13.71 & 0.535 & 0.380 & 24.75 & 0.845 & 0.129  \\
DepthSplat~\cite{xu2024depthsplat}  & 19.55 & 0.715 & 0.197 & 25.35 & 0.852 & 0.117  \\
LVSM~\cite{jin2025lvsm}             & 20.58 & 0.744 & 0.195 & 24.05 & 0.820 & 0.120  \\
Ours                       & \textbf{23.16} & \textbf{0.811} & \textbf{0.112} & \textbf{27.31} & \textbf{0.900} & \textbf{0.072}  \\
\bottomrule
\end{tabular}

\vspace{-4mm}
\label{tab:comparison_camnum}
\end{table}

%% file: tables/ablation_modules.tex
\begin{table}
\small
\vspace{-8mm}
\caption{\textbf{Quantitative ablation study of different modules.}
We evaluate the impact of different modules on performance.}
\centering
\setlength{\tabcolsep}{2.5pt}
\begin{tabular}{l|ccccc}
\toprule

\multicolumn{1}{c}{Model} & PSNR$\uparrow$ & SSIM$\uparrow$ & LPIPS$\downarrow$ & GS-Num & Time \\

\midrule

Full Model                & \textbf{27.18} & \textbf{0.891} & \textbf{0.071} & \textbf{71k} & \textbf{0.4} \\
\textit{w/o} Voxel Encode& 26.60 & 0.876 & 0.091 & 160k & 0.7 \\
\textit{w/o} Ray Casting& 26.32 & 0.856 & 0.093 & 224k & 0.7 \\

\bottomrule
\end{tabular}


\label{tab:ablation_modules}
\end{table}

%% file: tables/ablation_arch_geometry.tex







\begin{table}[h]
\small
\centering

\begin{minipage}{0.47\linewidth}
\centering
\caption{\textbf{Quantitative ablation study of different network architectures.}
We evaluate the impact of model architectures on performance. "AA" indicates the designed Alternating Attention.}
\setlength{\tabcolsep}{0.55pt}
\begin{tabular}{l|cccc}
\toprule
\multicolumn{1}{c}{Model} & PSNR$\uparrow$ & SSIM$\uparrow$ & LPIPS$\downarrow$ & Time \\
\midrule
AA(Ours) & \textbf{27.18} & \textbf{0.891} & \textbf{0.071} & \textbf{0.4} \\
Full & 27.09 & 0.887 & 0.076 & 1.1 \\
\bottomrule
\end{tabular}
\label{tab:ablation_arch}
\end{minipage}
\hfill
\begin{minipage}{0.47\linewidth}
\centering
\caption{\textbf{Quantitative ablation study of geometry proxies.}
We evaluate the impact of different geometry proxies on performance.}
\setlength{\tabcolsep}{0.55pt}
\begin{tabular}{l|ccc}
\toprule

\multicolumn{1}{c}{Model} & PSNR$\uparrow$ & SSIM$\uparrow$ & LPIPS$\downarrow$ \\

\midrule

Visual Hull(Ours)  & \textbf{27.18} & \textbf{0.891} & \textbf{0.071} \\
VGGT Depth   & 25.95 & 0.871 & 0.077 \\
Bounding Box & 23.89 & 0.775 & 0.153 \\
Frustum Sample & 22.92 & 0.749 & 0.184 \\

\bottomrule
\end{tabular}


\label{tab:ablation_geometry}
\end{minipage}
\end{table}

%% file: tables/ablation_voxel_size.tex
\begin{table}
\small
\vspace{-8mm}
\caption{\textbf{Quantitative ablation study on voxel size.}
We evaluate how voxel size affects quality–efficiency trade-offs.}
\centering
\setlength{\tabcolsep}{4.5pt}
\begin{tabular}{c|ccccc}
\toprule

\multicolumn{1}{c}{Voxel Size} & PSNR$\uparrow$ & SSIM$\uparrow$ & LPIPS$\downarrow$ & GS-Num & Time \\

\midrule

0.0025  & 27.18 & 0.894 & 0.074 & 238k & 1.4 \\
0.0030  & 27.52 & 0.897 & 0.072 & 189k & 0.9 \\
0.0050  & 27.18 & 0.891 & 0.071 & 71k & 0.4 \\
0.0100  & 25.51 & 0.853 & 0.102 & 30k & 0.3 \\

\bottomrule
\end{tabular}

\vspace{-3mm}

\label{tab:ablation_voxel_size}
\end{table}

%% file: sec/5_conclusion.tex
\section{Discussion}
\label{sec:discussion}

\paragraph{Conclusion.}
We presented \sysname{}, an object-centric, feed-forward framework that reconstructs compact 3D Gaussian representations directly from sparse RGB inputs.
By establishing appearance and geometry correspondences and introducing a Point-Image Transformer with modality-aware encodings,
our method jointly reasons about geometry and appearance in 3D space, avoiding the inter-view redundancy of view-centric pipelines.
This design scales efficiently with the number of views and image resolution, and supports real-time rendering from arbitrary viewpoints with lightweight user-side computation.

\paragraph{Limitations.}

Although our feed-forward method demonstrates significant advantages in efficiency and quality for novel view synthesis,
it may struggle with unbounded scenes due to memory constraints.
Additionally, extending it to construct temporally consistent and compact 4D representations remains an open challenge.
We believe that more memory-efficient architectures and 3D/4D representations could help address these limitations in future work.

%% file: sec/acknowledgements.tex
\section*{Acknowledgements}
This work was partially supported by National Key R\&D Program of China (No. 2024YFB2809105), Zhejiang Provincial Natural Science Foundation of China (No. LR25F020003), and Information Technology Center and State Key Lab of CAD\&CG, Zhejiang University.

%% file: sec/X_suppl.tex

\section{Method Details}

\PAR{Details of the Ray-Casting module.}
\label{sec:voxel-ray-query}
We voxelize the point cloud and traverse the grid with a vectorized DDA, allowing rays to run in parallel while querying candidate points in each voxel and testing their distances to the ray.
The points are sorted by voxel ID and indexed through compact offset/count tables, enabling each ray to scan only its valid box segment,
gather candidates along the traversal path, and continually update the closest hit.
The process terminates once the ray exits the bounding box, exceeds the current best hit, or reaches the step limit, with an optional pruning stage to remove rare duplicate voxel visits.
Pseudocode is provided in Algorithm~\ref{alg:ray-query}.


\begin{algorithm}[h]
\caption{Vectorized Voxel DDA Ray Query}
\label{alg:ray-query}
\begin{algorithmic}[1]
\Require Points $\{p_i\}$, rays $(o_j,d_j)$, voxel size $h$, radius $\varepsilon$
\Ensure Nearest hit per ray (none if no hit)

\State Compute AABB and grid from $h$; voxelize points, encode cell ids, sort; build dense tables \texttt{cell\_offset}, \texttt{cell\_count}.

\For{each ray $(o,d)$}
    \State Normalize $d$; intersect AABB to get $(t_{\min},t_{\max})$; \If{$t_{\max} < \max(t_{\min},0)$} \textbf{continue} \EndIf
    \State $t\gets\max(t_{\min},0)$; $g\gets\text{voxel}(o+td)$; $s\gets\text{sign}(d)$; precompute $t_{\Delta}$ and initial $t_{\text{max}}$; set best $(t^\star,p^\star)\gets(\infty,\text{none})$

    \For{$k=1$ to max\_steps}
        \If{$t>t_{\max}$ or $t>t^\star$} \textbf{break} \EndIf
        \State $(\text{off},\text{cnt}) \gets \text{lookup}(\texttt{cell\_offset},\texttt{cell\_count}, g)$
        \For{each $p$ in range $[\text{off},\text{off}+\text{cnt})$}
            \State $v\gets p-o$, $\hat{t}\gets v\cdot d$, $\delta\gets\|v-\hat{t}d\|$; \If{$\hat{t}\ge0\ \wedge\ \delta<\varepsilon\ \wedge\ \hat{t}<t^\star$} $(t^\star,p^\star)\gets(\hat{t},p)$ \EndIf
        \EndFor
        \State $a\gets\arg\min t_{\text{max}}$; $t\gets t_{\text{max}}[a]$; $t_{\text{max}}[a]\gets t_{\text{max}}[a]+t_{\Delta}[a]$; $g[a]\gets g[a]+s[a]$
    \EndFor

    \State Output $p^\star$
\EndFor
\end{algorithmic}
\end{algorithm}

\PAR{Details of the Point Encoding module.}
\label{sec:point-encoding}
We reuse the voxelization and hash table construction from the ray-casting module to group points within occupied voxels.
Each group is then converted into a fixed-size patch: if it contains more than $K$ points, we randomly sample $K$ of them; if it contains fewer, we repeat points to reach the target size.
Then, we compute the geometric center of each patch and encode it using sinusoidal projections defined over a frequency basis.
Finally, we concatenate the positional encoding with the point attributes and pass the result through an MLP to obtain a unified patch embedding.

\PAR{Details of the Point-Image Transformer.}
\label{sec:point-image-transformer-details}
The Point-Image Transformer comprises four blocks, each with three types of attention layers.
Each multi-head self-attention module employs 16 heads with RMS normalization applied to both query and key vectors.
We apply layer normalization prior to each attention and MLP layer, followed by residual connections after each block.
The feed-forward MLPs within each transformer block consist of two hidden layers with GELU activation functions.
The model contains approximately $190$M trainable parameters in total.

\section{Additional Results}
\input{tables/comparison_32views.tex}

\begin{figure}[t]
    \centering
    \includegraphics[width=0.99\linewidth]{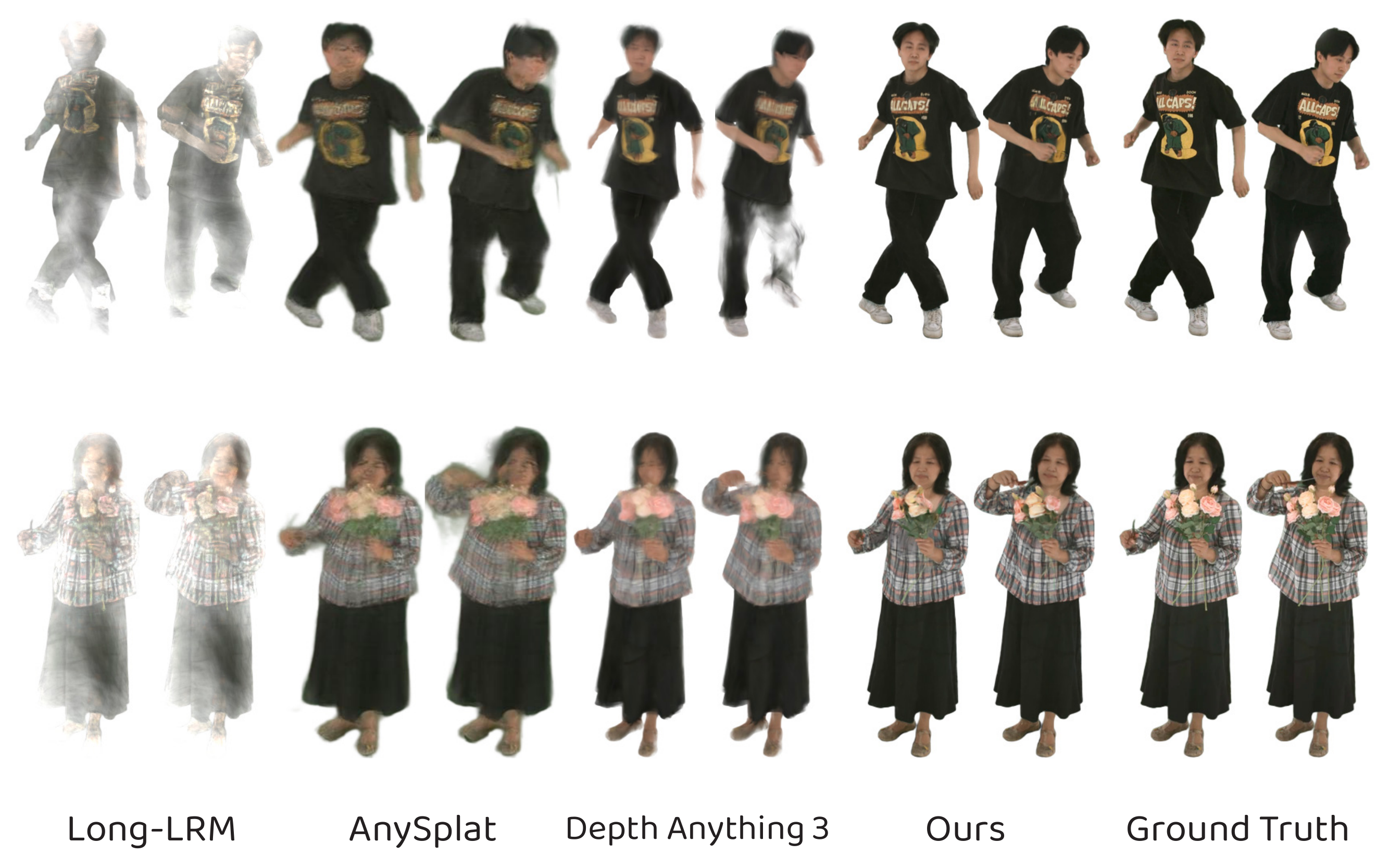}
    \vspace{-0.1in}
    \caption{\textbf{Qualitative comparison of dense input views}.
    Long-LRM's opacity-based pruning introduces ambiguity in learning per-view opacity weights, leading to uniform low-opacity distributions and resulting in transparent renderings.
    AnySplat prunes Gaussians in overlapping regions via voxelization, but geometric inaccuracies from its foundation model~\cite{wang2025vggt} cause artifacts.
    Depth Anything 3 reduces redundancy with confidence-based pruning, but struggles with high-frequency details, leading to lower rendering quality.
}
    \vspace{-0.2in}
    \label{fig:comparison_dense}
\end{figure}

\PAR{Results on dense input views.}
We compare our method with other methods designed for dense input settings with 32 views.
These methods typically compress the 3D Gaussian representation at the model's output stage, treating it as a post-processing step.
We consider three representative post-processing methods, including Depth Anything 3~\cite{depthanything3}, AnySplat~\cite{jiang2025anysplat}, and Long-LRM~\cite{ziwen2025llrm}.
As shown in Table~\ref{tab:comparison_dense} and Figure~\ref{fig:comparison_dense},
although these methods can reduce the number of Gaussians, they lead to significant degradation in rendering quality compared to our object-centric prediction.

\begin{figure*}
    \centering
    \includegraphics[width=0.8\linewidth]{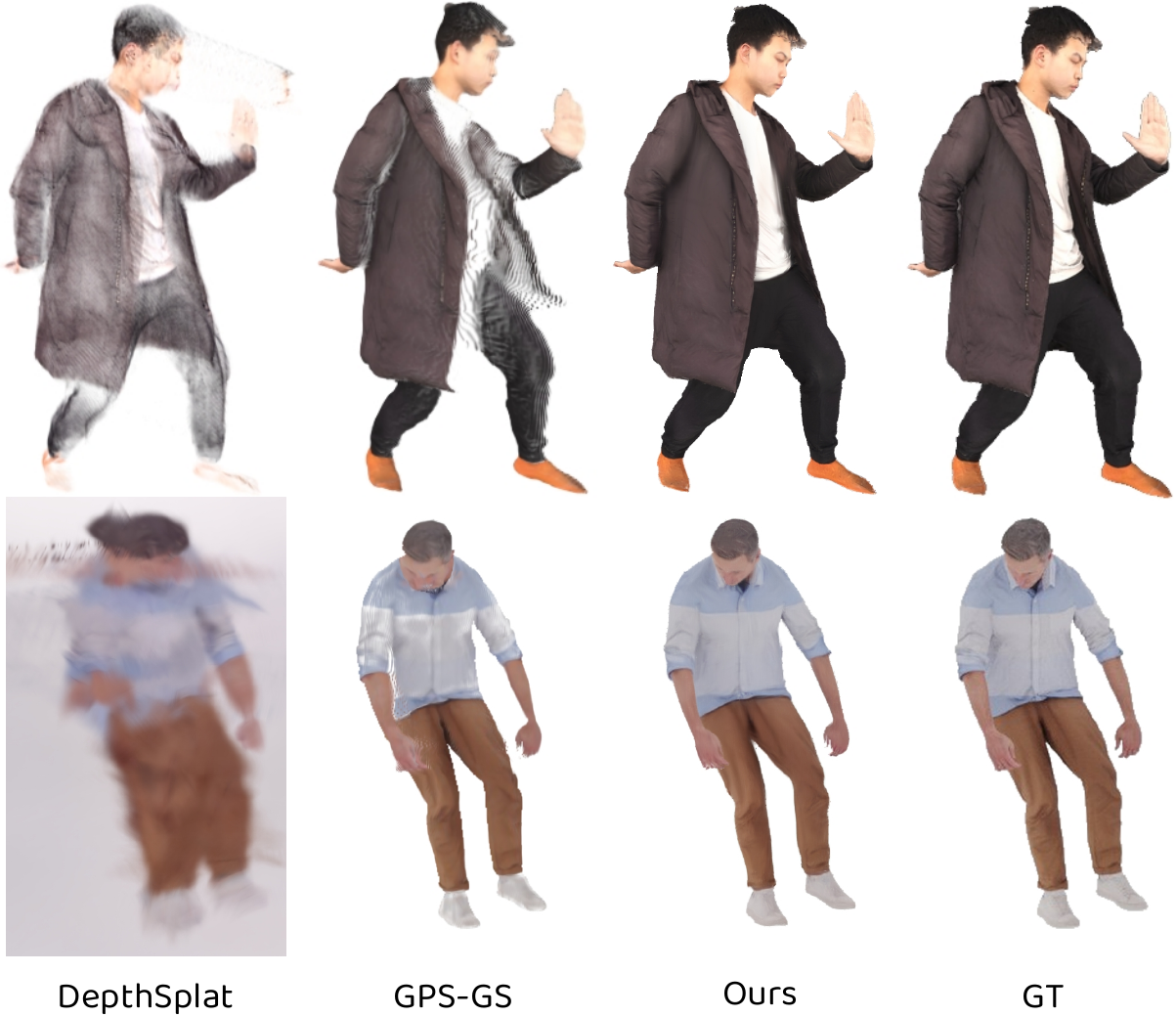}
    \vspace{-0.05in}
    \caption{\textbf{Results on synthetic datasets.}
    Additional qualitative examples on synthetic scenes.
    Our method produces sharp renderings with stable geometry and consistent textures under viewpoint changes.}
    \vspace{-0.1in}
    \label{fig:synthetic_results}
\end{figure*}

\PAR{Results on synthetic datasets.}
We provide additional qualitative results on synthetic scenes in Figure~\ref{fig:synthetic_results}.
The visualization shows that our method preserves fine structures and appearance consistency across novel viewpoints while maintaining compact Gaussian representations.

\begin{figure}
    \centering
    \includegraphics[width=0.6\linewidth]{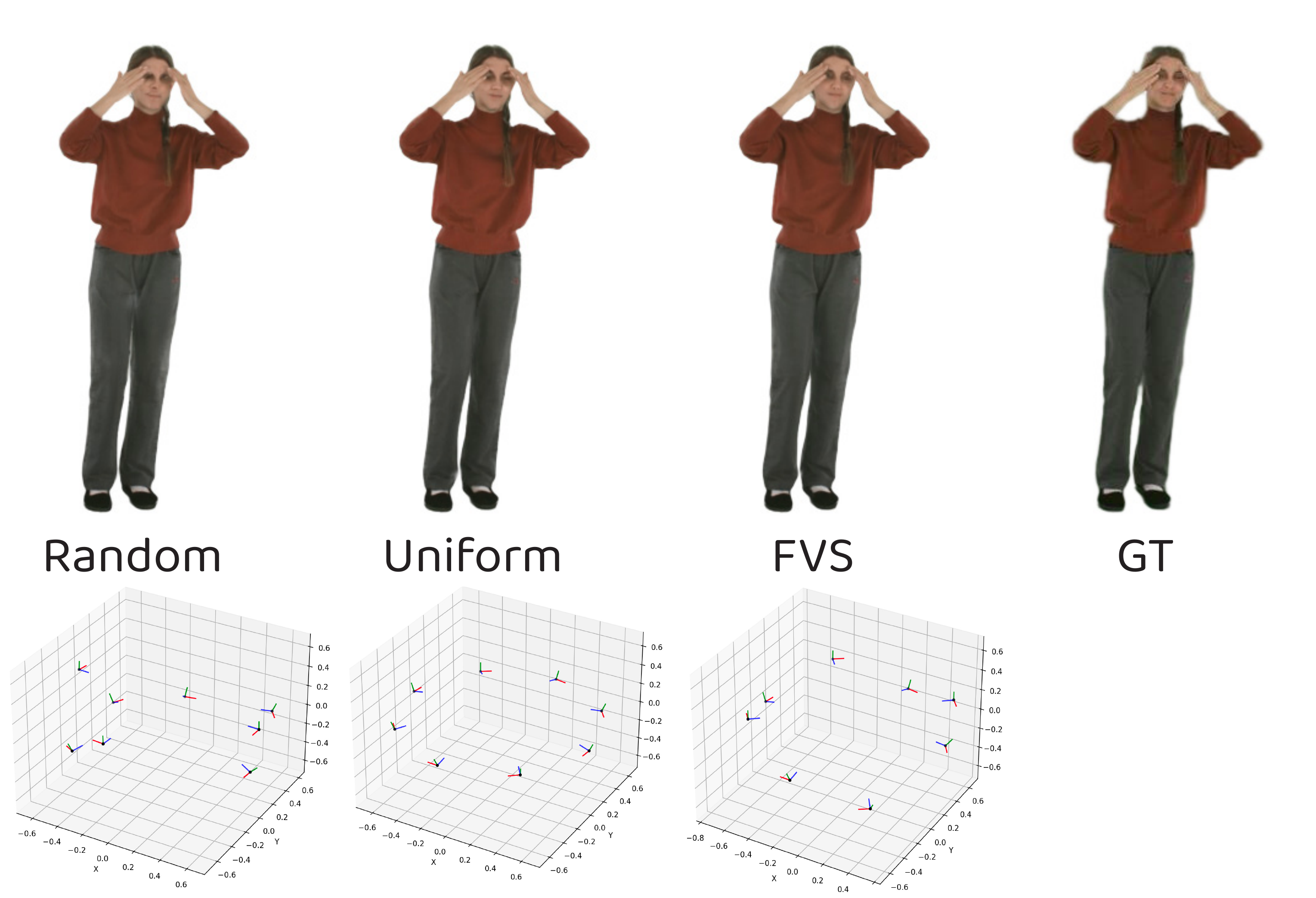}
    \vspace{-0.0in}
    \caption{\textbf{Results under different camera input setups.}
“Random” indicates randomly sampled views.
“Uniform” indicates uniformly sampled views from the middle column of the camera array in DNA-Rendering.
    “FVS” indicates the farthest-view sampling strategy used for input selection in training.}
    \vspace{-0.1in}
    \label{fig:different_setups}
\end{figure}

\PAR{Results on different camera setups.}
We train our model using views sampled via farthest view sampling from the DNA-Rendering dataset.
To evaluate the model's robustness, we test it with different camera input distributions during inference.
As shown in Figure~\ref{fig:different_setups},
our method predicts comparable results under these different camera setups, demonstrating strong generalization.

\begin{figure}[ht]
    \centering
    \includegraphics[width=0.8\linewidth]{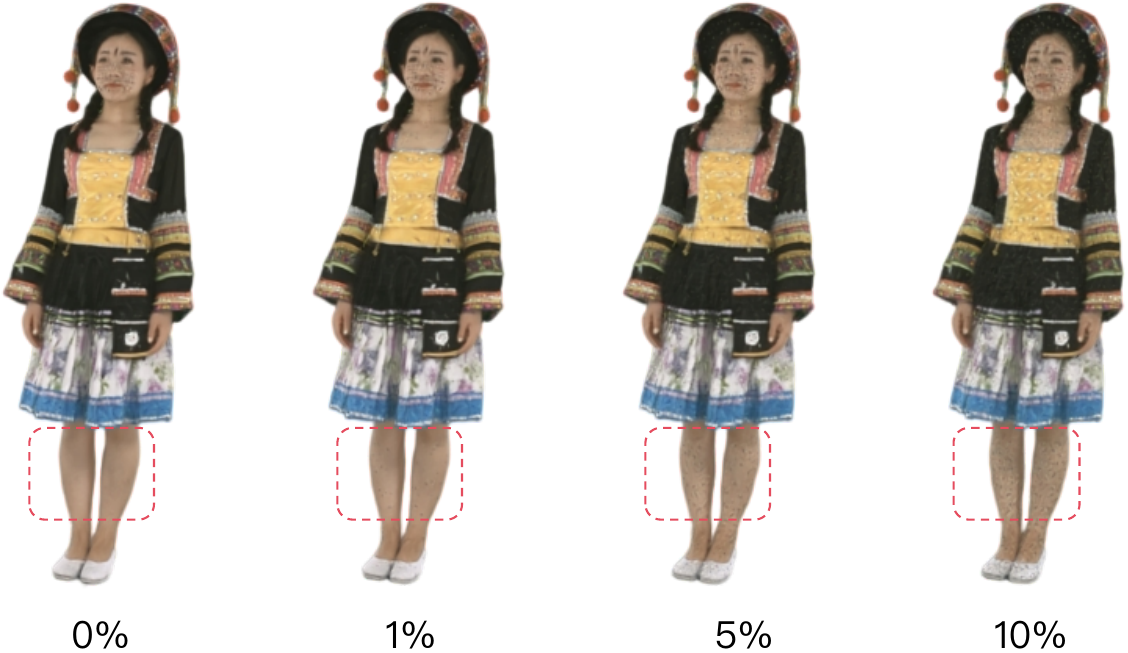}
    \caption{\textbf{Mask robustness under random foreground-pixel dropping.}
    The model remains robust under moderate mask corruption.}
    \label{fig:mask_robustness}
\end{figure}

\PAR{Mask robustness.}
For DNA-Rendering, we follow Diffuman4D~\cite{jin2025diffuman4d} and use preprocessed masks for both training and evaluation, since the provided masks may contain incomplete foreground regions.
All baselines use the same masks to ensure fair input conditions.
To evaluate robustness to mask noise, we randomly drop foreground mask pixels at different ratios during inference.
As shown in Table~\ref{tab:mask_robustness} and Figure~\ref{fig:mask_robustness}, the proposed method degrades gracefully and maintains reasonable rendering quality even at a 10\% drop ratio.
\begin{table}[ht]
\small
\caption{\textbf{Quantitative mask robustness under random foreground-pixel dropping.}}
\centering
\setlength{\tabcolsep}{4pt}
\begin{tabular}{c|ccc}
\toprule

Drop ratio & PSNR$\uparrow$ & SSIM$\uparrow$ & LPIPS$\downarrow$ \\

\midrule

0\%  & 26.74 & 0.878 & 0.087 \\
1\%  & 26.52 & 0.859 & 0.145 \\
5\%  & 25.66 & 0.796 & 0.214 \\
10\% & 24.55 & 0.743 & 0.253 \\

\bottomrule
\end{tabular}
\label{tab:mask_robustness}
\end{table}

\begin{figure}
    \centering
    \vspace{-0.1in}
    \includegraphics[width=0.58\linewidth]{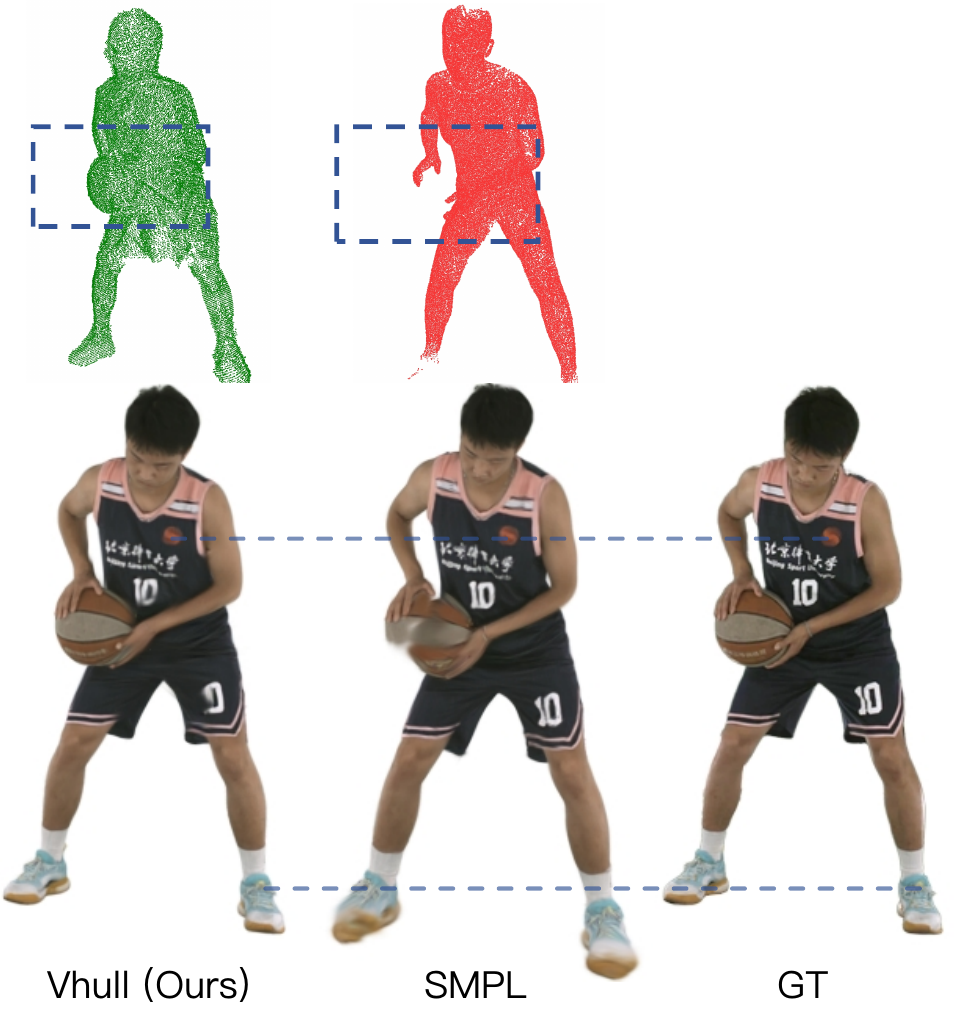}
    \caption{\textbf{Comparison of SMPL and visual-hull initialization.}
    Visual-hull anchors better cover non-template geometry such as loose clothing and objects.}
    \label{fig:smpl_vs_vhull}
    \vspace{-0.2in}
\end{figure}

\PAR{Initialization with visual hulls and SMPL.}
We compare visual-hull and SMPL-based initialization in Figure~\ref{fig:smpl_vs_vhull}.
SMPL-based points are constrained by mocap accuracy and the template geometry, making them less reliable for loose clothing and human-object interactions.
In contrast, visual-hull anchors better cover the observed foreground geometry and provide a more flexible initialization for our point-set prediction.

\begin{figure}
    \centering
    \vspace{-0.3in}
    \begin{subfigure}{0.38\linewidth}
        \centering
        \includegraphics[width=\linewidth]{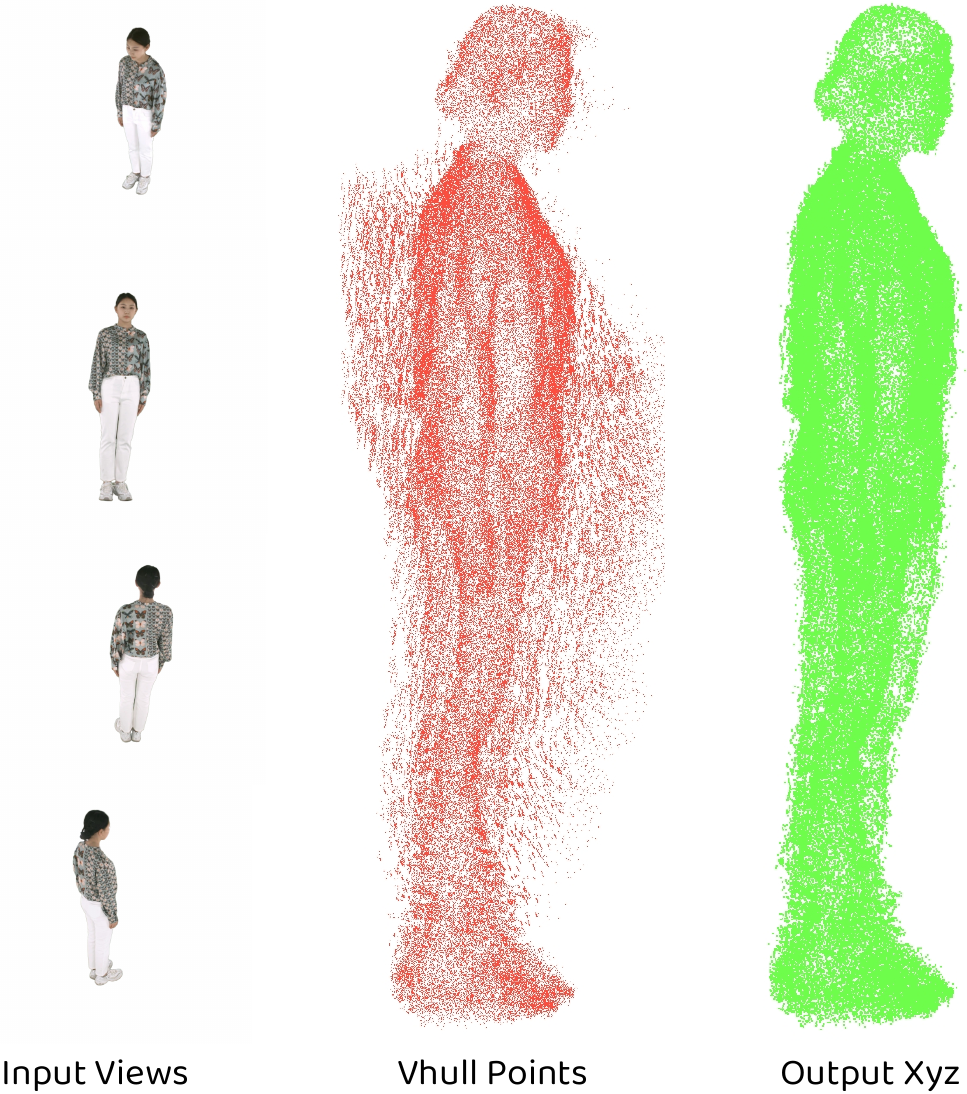}
        \caption{\textbf{Point-set offsets.}
        Offset regression corrects noisy visual-hull anchors.}
        \label{fig:point_offset}
    \end{subfigure}
    \hfill
    \begin{subfigure}{0.44\linewidth}
        \centering
        \includegraphics[width=\linewidth]{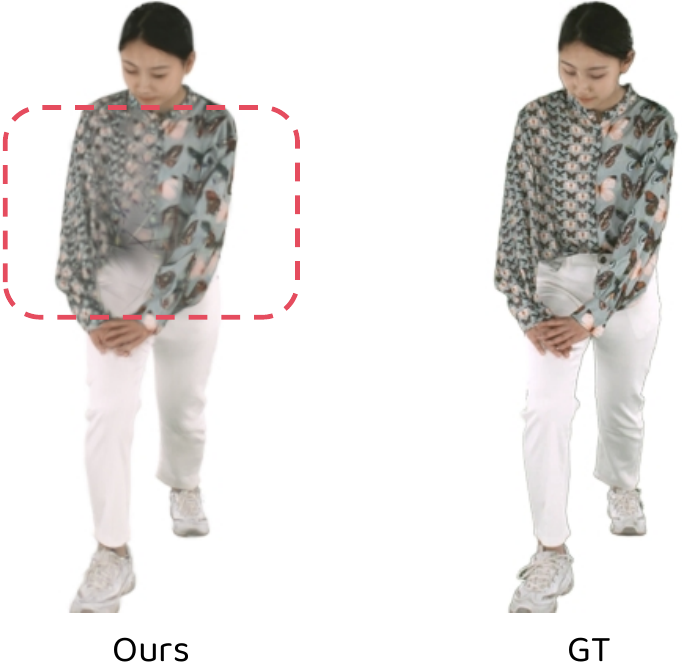}
        \caption{\textbf{Failure case.}
        Severe self-occlusion leads to blurred reconstruction.}
        \label{fig:failure_cases}
    \end{subfigure}
    \caption{\textbf{Additional analysis of point-set prediction.}
    Left: predicted offsets recover accurate geometry from noisy visual-hull anchors.
    Right: the method may produce blurred results when large regions are occluded.}
    \vspace{-0.3in}
\end{figure}

\PAR{Visualization of the predicted point-set offsets.}
The visual hull is only used to provide query anchors, while our network regresses offsets relative to these anchors.
This design makes the model less sensitive to visual-hull quality and moderate calibration noise.
As shown in Figure~\ref{fig:point_offset}, even when the hull is noisy in sparse-view settings, the predicted offsets can correct geometric errors and recover accurate geometry.
We also observe limited sensitivity to the exact camera layout, consistent with the results in Figure~\ref{fig:different_setups}.

\PAR{Failure cases.} 
Under severe self-occlusion where large regions are not visible, the reconstruction becomes blurred; an example is shown in Figure~\ref{fig:failure_cases}.

%% file: tables/comparison_32views.tex
\begin{table}[!t]
\small
\caption{\textbf{Quantitative comparison on dense input views.}
We evaluate the zero-shot generalization of our method to 32 input views.
Our method remains robust as the number of input views increases.
}
\centering
\setlength{\tabcolsep}{2pt}
\begin{tabular}{l|ccccc}
\toprule

\multicolumn{1}{c}{Model} & PSNR$\uparrow$ & SSIM$\uparrow$ & LPIPS$\downarrow$ & GS-Num & Time \\

\midrule

Ours                     & \textbf{25.46} & \textbf{0.854} & \textbf{0.087} & \textbf{76k} & \textbf{1.3} \\
DA3~\cite{depthanything3}  & 20.48 & 0.690 & 0.231 & 666k & 5.4 \\
AnySplat~\cite{jiang2025anysplat} & 17.38 & 0.650 & 0.305 & 751k & 3.3 \\
Long-LRM~\cite{ziwen2025llrm}       & 15.74 & 0.640 & 0.304 & 423k & 1.6 \\

\bottomrule
\end{tabular}

\vspace{-3mm}
\label{tab:comparison_dense}
\end{table}